%% file: fast_grl_iclr_workshop.tex
\documentclass{article}
\usepackage{iclr2021_workshop,times}

\input{math_commands.tex}

\usepackage{changepage}
\usepackage{booktabs}
\usepackage{algorithm}
\usepackage[noend]{algpseudocode}
\usepackage{xfrac}
\usepackage{arydshln}
\usepackage{mathtools}
\usepackage{amsfonts}
\usepackage{enumitem}
\usepackage{amsmath}
\usepackage{amssymb}
\usepackage{xcolor}

\newtheorem{theorem}{Theorem}

\makeatletter
  \renewcommand*\env@matrix[1][*\c@MaxMatrixCols c]{%
    \hskip -\arraycolsep
    \let\@ifnextchar\new@ifnextchar
  \array{#1}}
\makeatother

\usepackage{hyperref}
\usepackage{url}

\title{Fast Graph Learning \\ with Unique Optimal Solutions}

\author{Sami Abu-El-Haija, Valentino Crespi, Greg Ver Steeg, Aram Galstyan \\
USC Information Sciences Institute \\
\texttt{\{\href{mailto:haija@isi.edu}{haija}, \href{mailto:vcrespi@isi.edu}{vcrespi}, \href{mailto:gregv@isi.edu}{gregv}, \href{mailto:galstyan@isi.edu}{galstyan}\}@isi.edu}
}

\iclrfinalcopy % Uncomment for camera-ready version, but NOT for submission.
\begin{document}

\maketitle

\begin{abstract}
We consider two popular Graph Representation Learning (GRL) methods: message passing for node classification and network embedding for link prediction.
For each, we pick a popular model that we:
(i) \textit{linearize} and (ii) and switch its training objective to \textit{Frobenius norm error minimization}.
These simplifications can cast the training into finding
the optimal parameters in closed-form.
We program in TensorFlow a functional form of
Truncated Singular Value Decomposition (SVD), such that, we could decompose a dense matrix $\mathbf{M}$, without explicitly computing $\mathbf{M}$.
We achieve competitive performance on popular GRL tasks while providing orders of magnitude speedup. We open-source our code at
\url{http://github.com/samihaija/tf-fsvd}
\end{abstract}

\section{Introduction}
\label{intro}
%{\bf AG: The intro should follow the strcuture of the abstract, e.g., dont start with SVD, but motivate the need for quick and accurate baselines} 

%Nowadays, many advancements in graph representation learning (GRL)  are supported by powerful computational frameworks such as TensorFlow \citep{tensorflow} and PyTorch \citep{pytorch}.
%Implementing popular transformations, such as neural layers for composing deeper networks,
Many recent  graph representation learning (GRL) models
are creative and theoretically-justified
\citep{kipf, sage, gat, qiu2018network, GIN, mixhop, GCNII}.
Unfortunately, however, they contain hyperparameters that need to be tuned (such as learning rate, regularization coefficient, depth and width of the network), and training takes a long time (e.g. minutes) even on smaller datasets.
%: the time only increases with the graph size.
%Further, it is rather uncommon to check for convergence, e.g., by checking $\nabla_\mathbf{W} J = 0$ where $J$ is mean training error.

We circumvent these weaknesses.
%Our main goal is to effectively train simple yet powerful GRL models and baselines.
%In general, these applications of SVD should  all benefit from an efficient implementation for structered data.
%In this paper, we are broadly interested in two
%GRL tasks: node classification and link prediction. 
We
(i) \textbf{quickly} train (ii) \textbf{competitive} GRL models by posing convex objectives and estimating optimal solutions in
closed-form, hence
(iii) \textbf{relieving} practitioners from hyperparameter tuning or convergence checks.
Our goals remind us of a \textit{classical learning technique} that has been used for decades.
Specifically, 
Singlar Value Decomposition
(SVD).

SVD periodically
appears
within powerful yet simple methods, competing on state-of-the-art.
The common practice is
to design a matrix $\mathbf{M}$, such that its decomposition (via SVD), provides an estimate for learning a model given an objective. 
For instance, \citet{levy2014-neural}
show that the learning of NLP skipgram models such as word2vec \citep{word2vec} and GloVe \citep{glove}, can be approximated by the SVD of a \textit{Shifted Positive Pointwise Mutual Information} matrix.

%This arguably motivated \citet{qiu2018network}
%to use SVD for approximating graph embedding methods, including the models of \citet{perozzi2014deepwalk, tang2015line, grover2016node2vec}.

In GRL, \citet{ultimatewalk, qiu2018network, wys}  have approximated methods of DeepWalk \citep{perozzi2014deepwalk} and Node2Vec \citep{grover2016node2vec} via decomposition of some matrix $\mathbf{M}$. However, their decomposition requires $\mathbf{M}$ to be either (a) exactly calculated or (b) sampled entry-wise, but (a) is unnecessarily expensive for real-world large networks (due to Small World Phenomenon, \citep{travers1969study}) and (b) incurs unnecessary estimation errors.
%
%Unfortunately,
%in cases of large graphs,
%$\mathbf{M}$ could hold higher-order information, making it occupy quadratic space in the number of graph nodes.
On the other hand, known algorithms in matrix theory can decompose \textit{any} matrix $\mathbf{M}$ \textit{without explicitly knowing}  $\mathbf{M}$. Specifically, it is sufficient to provide a  function $f_\mathbf{M}(.) = \langle \mathbf{M}, . \rangle$ that can multiply $\mathbf{M}$ with arbitrary vectors (§\ref{sec:fsvd}).
We, argue that if the popular frameworks (e.g., TensorFlow) implement a \textit{functional SVD}, that accept $f_\mathbf{M}(.)$ rather than $\mathbf{M}$, then modern practitioners may find it useful.

%Although a functional SVD could assist various fields, we focus our interest on GRL models, specifically ones that can output predictions per node  (e.g., for node classification) or a per edge (e.g., for link prediction).
%GRL can benefit from a
%functional SVD, as many methods can be approximated by SVD of
%some matrix
%$\mathbf{M} = \sum_i g_i(\mathbf{A})^i$, where adjacency $\mathbf{A}$ and its transformations $g_i$ are all sparse matrices, though the power $i \in \mathbb{N}_+$ likely yields dense $g_i(\mathbf{A})^i$ for $i \ge 6$ hence requiring quadratic storage %(§\ref{sec:smallworld}).
%In these cases, even though $\mathbf{M}$ is expensive to calculate, it is much cheaper to multiply it by vectors, therefore performing learning using memory and computation complexity that is linear in network size, while sacrificing no more error than SVD on dense $\mathbf{M}$.

%
We review powerful GRL methods (§\ref{sec:prelim_grl}) that we \textit{convexify} (§\ref{sec:cvx}),
allowing us to use
% by following two steps: (i) linearize their model and (ii) replace their objective with Frobenius norm minimization.
(randomized) SVD for obtaining
(approximate) optimum solutions.
Our contributions are:
\begin{enumerate}[itemsep=0pt, topsep=0pt, leftmargin=12pt]
\item We implement
a functional SVD (\S\ref{sec:fsvd})  of the randomized algorithm of \citet{halko2009svd}.

\item We approximate embedding and message passing methods via SVD (\S\ref{sec:cvx}), showing competitive performance with state-of-the-art,
yet much faster to train (\S\ref{sec:exp}).
\item We analyze that learning is fast and approximation error can be made arbitrarily small (\S\ref{sec:analysis}).
\end{enumerate}

\section{Preliminaries}
\subsection{Singular Value Decomposition (SVD)}

%The $k$-truncated SVD of an $m\times n$ matrix $M$ is the matrix
 % $M_k=U_k\Sigma_k V_k^T$ where $\Sigma_k$ is a diagonal matrix containing the top $k$ singular values while $U_k$ and $V_k^T$ are unitary matrices of dimensions $m\times k$ and $k\times m$ respectively. 

Truncated (top-$k$) Singular Value Decomposition (SVD) estimates input matrix
%a real\footnote{Our algorithm readily applies to complex$\mathbf{M} \in \mathbb{C}^{r \times c}$, swapping $\mathbf{V}^\top$ with $\mathbf{V}^*$ everywhere}
$\mathbf{M} \in \mathbb{R}^{r \times c}$ with  low-rank estimate  ${\widetilde{\mathbf{M}}}$ that minimizes the Frobenius norm of the error:
%and rank $k \in \mathbb{N}_{\ge 1}$, and finds then solves:
\begin{equation}
\min_{\widetilde{\mathbf{M}}} ||\mathbf{M} - {\widetilde{\mathbf{M}}}||_F^2, \textrm{ \ \ \ \  subject to: \ \ } \textrm{rank}(\widetilde{\mathbf{M}}) \le k,
\end{equation}
while parameterizing $\widetilde{\mathbf{M}}$ as $\widetilde{\mathbf{M}} = \mathbf{U}_k \mathbf{S}_k \mathbf{V}_k^\top$, subject to, columns of $\mathbf{U}_k, \mathbf{V}_k$ being orthonormal.
It turns out,
the minimizer of
Frobenius norm $||.||_F$
recovers
the  top-$k$ singular values  (stored along diagonal matrix $\mathbf{S}_k \in \mathbb{R}^{k \times k}$), with their corresponding left- and right-singular vectors, respectively, stored as columns of the unitary matrices $\mathbf{U}_k \in \mathbb{R}^{r \times k}$ and $\mathbf{V}_k \in \mathbb{R}^{c \times k}$ (\textit{a.k.a}, the \textit{singular bases}).

%where unitary matrices $\mathbf{U}_k \in \mathbb{R}^{r \times k}$ and $\mathbf{V}_k \in \mathbb{R}^{c \times k}$ form a \textit{bases}, and diagonal matrix $\mathbf{S}_k$  contains the top-$k$ singular values, such that, $\widetilde{\mathbf{M}} = \mathbf{U}_k \mathbf{S}_k \mathbf{V}_k^\top$ 
% Mention pseudoinverse application?
%
%There are a number of SVD algorithms for estimating $\widetilde{\mathbf{M}}$, such as the variants of the power iteration method with orthogonal projections \citep{lanczos1950iteration, arpack}. In this paper, we provide a \textit{functional} abstraction of the randomized SVD algorithm of \citet{halko2009svd}, which we describe in §\ref{sec:fsvd}.
SVD has many applications and we utilize two:
(1)
it is used for embedding and matrix completion; and (2) it can estimate the pseudoinverse of matrix $\mathbf{M}$ (a.k.a., Moore-Penrose inverse), as:
\begin{equation}
\label{eq:pinv_approx_svd}
\mathbf{M}^\dagger \triangleq {\mathbf{M}}^\top \left({\mathbf{M}} {\mathbf{M}}^\top \right)^{-1}   \approx \mathbf{V}_k \mathbf{S}_k^{-1} \mathbf{U}_k^\top,
\end{equation}
where one calculates inverse $\mathbf{S}^{-1}$ by  reciprocating entries of diagonal matrix $\mathbf{S}$. The $\approx$ becomes $=$ when $k \ge \textrm{rank}(\mathbf{M})$, due to \citep{eckart1936lowrank, golub1996matrix}.

\subsection{Graph Representation Learning (GRL)}
\label{sec:prelim_grl}
%We review two popular GRL methods and tasks:
%(i) \textit{message passing} for node classification and (ii) %\textit{network embedding} for link prediction, as they cover many %practical applications including in recommender systems, biological e.g. protein-protein interactions, social and citation networks:
\begin{itemize}[itemsep=0pt, topsep=0pt,leftmargin=12pt]
\item[(i)] Many \textit{message passing} models can be written as:
\begin{equation}
\label{eq:prelim_mp}
\mathbf{H} = \sigma_L\Bigg(g_L(\mathbf{A}) \dots \ \  \overbrace{\sigma_2\bigg(g_2(\mathbf{A}) \ \ \underbrace{\sigma_1 \left( g_1(\mathbf{A}) \mathbf{X} \mathbf{W}_1 \right)}_{\textrm{output of layer 1}} \mathbf{W}_2 \bigg)}^{\textrm{output of layer 2}} \dots \mathbf{W}_L \Bigg)
\end{equation}
%\\
where $L$ is the number of layers, matrix $\mathbf{X}\in \mathbb{R}^{n \times d}$ contains $d$ features per node, $\mathbf{W}$'s are trainable parameters, $\sigma$ denote activations (e.g. ReLu), and $g$ is some (possibly trainable) transformation of adjacency matrix. GCN \citep{kipf} set $g$ to symmetric normalization per \textit{renormalization trick}, GAT \citep{gat} set $g(\mathbf{A}) = \mathbf{A} \circ \textrm{MultiHeadedAttention}$ and GIN \citep{GIN} as $g(\mathbf{A}) = \mathbf{A} + (1+\epsilon) \mathbf{I} $ with identity $\mathbf{I}$ and $\epsilon > 0$.
For node classification, it is common to set $\sigma_L = \textrm{softmax}$ (applied row-wise), specify the size of $\mathbf{W}_L$ s.t. $\mathbf{H} \in \mathbb{R}^{n \times y}$ where $y$ is number of classes, and optimize cross-entropy objective:
\begin{equation}
\label{eq:xe_mp}
\min_{ \{\mathbf{W}_j\}_{j=1}^L}  -  \mathbf{Y} \circ \log \mathbf{H} -  (1-\mathbf{Y}) \circ \log (1-\mathbf{H}),
\end{equation}
where $\mathbf{Y}$ is a binary matrix with one-hot rows indicating node labels. $\circ$ is Hadamard product.
in semi-supervised node classification settings where not all nodes are labeled, before measuring the objective, subset of rows can be kept in $\mathbf{Y}$ and $\mathbf{H}$ that correspond to labeled nodes.
\item[(ii)] \textit{Network embedding} methods map nodes onto a $z$-dimensional vector space $\mathbf{Z} \in \mathbb{R}^{n \times z}$.
%Classical approaches compute $\mathbf{Z}$ by directly decomposing a transformation of the adjacency matrix, such as the graph laplacian \citep{eigenmaps}.
%The embeddings can then be used to solve graph tasks, such as link prediction or node classifications.
Modern approaches train skipgram models (e.g. word2vec \citep{word2vec}) on sampled random walks.  It has been shown that these skipgram network embedding methods, including DeepWalk \citep[][]{perozzi2014deepwalk}) and  node2vec \citep[][]{grover2016node2vec}, with a learning process of walk sampling followed by positional embedding, can be approximated as a matrix deomposition \citep{ultimatewalk, wys, qiu2018network}. We point the curious reader to the listed papers for how the decomposition was derived, but show here the derivation of \citet[WYS,][]{wys}, as it performs well in our experiments:
%\vspace{-2mm}
\begin{equation}
\label{eq:embedding}
%\textrm{\resizebox{.99\linewidth}{!}{$
\min_{\mathbf{Z}} - \mathbf{M}^{^\textrm{(WYS)}} \circ \log h(\mathbf{Z}) - (1 - \mathbf{A}) \circ \log (1 - h(\mathbf{Z})),
%$}}
\end{equation}
where $h(\mathbf{Z})=h([
\begin{matrix}[c;{1pt/1pt}c]
\mathbf{L} &  \mathbf{R} \end{matrix} ] ) =(1+\exp(\mathbf{L} \times \mathbf{R}^\top  ))^{-1}$ i.e. $\mathbf{Z}$ concatenates $\mathbf{L},  \mathbf{R} \in \mathbb{R}^{n \times \frac{z}{2}}$ and $h$ is the logistic of their cross-correlation (pairwise dot-products).
$\mathbf{M}^{^\textrm{(WYS)}} =\sum_i^C  (\mathbf{D}^{-1} \mathbf{A})^i \mathbf{c}_i   = \sum_{i=1}^C    \mathcal{T}^i \mathbf{c}_i$,
%\end{equation}
%\mathbf{D}
where $\mathcal{T}$ is the transition matrix, $\mathbf{D}=\textrm{diag}(\mathbf{1}^\top \mathbf{A})$ is diagonal degree matrix,  and we fix vector $\mathbf{c}$ to staircase: $\mathbf{c}_i = C-i+1$. For instance, $\mathbf{c} = [4, 3, 2, 1]$ for context size $C=4$.

\end{itemize}

%We define the goal of network embedding as follows. Given a graph with $n$ nodes and $m$ edges, an embedding algorithm constructs a vector space (so called, the embedding space) in which utilizes the edges 

%\citet{perozzi2014deepwalk} proposed a class methods for embedding large networks:
%Given a graph, these methods operate in two steps: (i) simulate a large number of random walks on the input graph, each walk resulting in a sequence node IDs and (ii) the node ID sequences are passed to an language embedding model, such as Skipgram of word2vec \citep{}.

\section{Our Proposed Convex Objectives}
\label{sec:cvx}
\subsection{Network Embedding Model}
\label{sec:embeddingmodel}
Objective in Eq~\ref{eq:embedding}
learns node embeddings $\mathbf{Z}=[\begin{matrix}[c;{1pt/1pt}c]
\mathbf{L} &  \mathbf{R} \end{matrix}] \in \mathbb{R}^{n \times z}$ using %(probabilistic)
cross-entropy. The terms: model output (outer product, $\sigma(\mathbf{L} \times \mathbf{R}^\top)$), negatives (non-edges, $1-\mathbf{A}$), and positives (expected number of node pairs covisits, $\mathbf{M}$), are all dense matrices with $\mathcal{O}(n^2) \gg m$ nonzero entries.
For instance, even a relatively-small social network with $n$=100,000 and average degree of 100 (i.e. $m=100n$) would produce an $\mathbf{M}$ occupying $\approx$40GB memory, whereas one can do the entire learning with $\approx$40MB memory using functional SVD (§\ref{sec:fsvd}).
%, specifically, by
We start by designing a matrix $\widehat{\mathbf{M}}^{^\textrm{(WYS)}}$
incorporating positive and negative information ($\mathbf{M}^{^\textrm{(WYS)}}$ and $1 - \mathbf{A}$) as:
%setting $\mathbf{Z}$ using the SVD \textit{basis} $(\mathbf{U}, \mathbf{S}, \mathbf{V})$. To do so, we design 
%that incorporates the \textit{positive} and \textit{negative} information as: %$\mathbf{M}^{^\textrm{(WYS)}}$ and $(1 - \mathbf{A})$, as:
\begin{align}
%\mathbf{M}^{^\textrm{(WYS)}} = 
\widehat{\mathbf{M}}^{^\textrm{(WYS)}} &= \mathbf{M}^{^\textrm{(WYS)}} - \lambda (1 - \mathbf{A}) = \sum_i    \mathcal{T}^i  \mathbf{c}_i - \lambda (1 - \mathbf{A}), \label{eq:m_wys}
\end{align}
with coefficient $\lambda \ge 0$ weighing negative samples. We use $\widehat{\ .\ }$ to denote our convexification.

%\subsubsection{Learning}
%\label{sec:embeddinglearning}
\textbf{Learning:} We can directly set $\mathbf{Z}$ to the SVD \textit{basis} $(\mathbf{U}, \mathbf{S}, \mathbf{V})$.
Matrix $\widehat{\mathbf{M}}^{^\textrm{(WYS)}}$ has  large entry $\widehat{\mathbf{M}}_{uv}$ when nodes ($u, v$) are well-connected (co-visited many times, during random walks) and small if they are non-edges.
%Suppose we have a low-rank estimator of
SVD provides a rank $k$ estimator of
$\widehat{\mathbf{M}}$ as $\widehat{\mathbf{L}}  \widehat{\mathbf{R}}^\top \approx \widehat{\mathbf{M}}$ i.e. with minimum Frobenius norm of error.
%Fortunately,  $\widehat{\mathbf{M}}$ by finding the rank-$k$ basis .
%large $\widehat{\mathbf{M}}_{uv}$ when $\mathbf{M}_{uv}$ is large, and small nodes $u$ and $v$ share no edge. Embeddings can now be trained with rank-$k$ SVD:
We can set the network embedding model parameters  $\mathbf{Z}=[\begin{matrix}[c;{1pt/1pt}c]
\widehat{\mathbf{L}} &  \widehat{\mathbf{R}} \end{matrix}]$ as:
%$$:
\begin{align}
\textrm{SVD}(\widehat{\mathbf{M}}, k) & \leftarrow  \argmin_{\widehat{\mathbf{U}}_k \widehat{\mathbf{S}}_k \widehat{\mathbf{V}}_k} 
\left|\left| \widehat{\mathbf{M}} - \widehat{\mathbf{U}}_k \widehat{\mathbf{S}}_k \widehat{\mathbf{V}}_k^\top \right|\right|_F^2 =\argmin_{\widehat{\mathbf{U}}_k \widehat{\mathbf{S}}_k \widehat{\mathbf{V}}_k} 
\left|\left| \widehat{\mathbf{M}} - \underbrace{\left(\widehat{\mathbf{U}}_k \widehat{\mathbf{S}}_k^\frac12\right)}_{\triangleq \widehat{\mathbf{L}} } \underbrace{\left(\widehat{\mathbf{S}}_k^\frac12 \widehat{\mathbf{V}}_k^\top\right)}_{\triangleq \widehat{\mathbf{R}}^\top } \right|\right|_F^2 
\end{align}
\begin{flalign}
\label{eq:compute_lr}
\textrm{Learning follows as:}
\hspace{0.6cm}
\widehat{\mathbf{U}}_k, \widehat{\mathbf{S}}_k, \widehat{\mathbf{V}}_k  \leftarrow \textrm{SVD}(\widehat{\mathbf{M}}^{^\textrm{(WYS)}}, k); \hspace{0.6cm}
\widehat{\mathbf{L}}  \leftarrow \widehat{\mathbf{U}}_k \widehat{\mathbf{S}}_k^{\frac12};
\hspace{0.6cm}
\widehat{\mathbf{R}}  \leftarrow \widehat{\mathbf{V}}_k \widehat{\mathbf{S}}_k^{\frac{1}{2}} &&
\end{flalign}

%\subsubsection{}
%\label{sec:embeddinginference}
\begin{flalign}
\textrm{\textbf{Inference:} Given query edges $Q=\{(u_i, v_i)\}_i$, one can compute model:} \hspace{0.1cm}
\mathbf{H}_Q = \widehat{\mathbf{R}}_{\{v\}_i}^\top \widehat{\mathbf{L}}_{\{u\}_i}, &&
\end{flalign}
where the (RHS) set-subscript denotes \textit{gathering} rows ({a.k.a, advanced indexing}), and $\mathbf{H}_Q \in \mathbb{R}^{|Q|}$.

%in one of the earliest 
%would amounts to learning multiplications in the , one of the first definitions of \textit{graph convolution}, ).   according   Although Even though it is possible to

\subsection{Messaging Passing Models}
\label{sec:classification_learning}
%\subsubsection{Model}
We can linearize \textbf{message passing} models (Eq.~\ref{eq:prelim_mp}) by assuming all $\sigma$'s are identity ($\sigma(.) = .)$.
%To simplify the presentation (though not necessary)
Let $g = g_1 = g_2 = \dots$, specifically let $g(\mathbf{A}) = (\mathbf{D} + \mathbf{I})^{\sfrac{-1}{2}} (\mathbf{A} + \mathbf{I}) (\mathbf{D} + \mathbf{I})^{\sfrac{-1}{2}}$ per renormalization trick of \citep{kipf}. A linear $L$-layer message passing network can be:
\begin{equation}
\label{eq:h_jkn}
\widehat{\mathbf{H}} = \underbrace{\left[
\begin{matrix}[c;{1pt/1pt}c;{1pt/1pt}c;{1pt/1pt}c;{1pt/1pt}c]
\mathbf{X} &  g(\mathbf{A}) \mathbf{X} & g(\mathbf{A})^2 \mathbf{X} & \dots & g(\mathbf{A})^L \mathbf{X} \end{matrix} \right]}_{\hspace{1.2cm} \overset{\Delta}{=} \ \  \widehat{\mathbf{M}}^{\textrm{(JKN)}} } \widehat{\mathbf{W}}.
\end{equation}
Concatenation of all layers was proposed in Jumping Knowledge Networks \citep[JKN,][]{jknet}.
%In this case, one can reinterpret $\widehat{\mathbf{W}}$ as row-wise concatenation of $L+1$ matrices:
%$\widehat{\mathbf{W}}^\top =\left[ \begin{matrix}[c;{1pt/1pt}c;{1pt/1pt}c] \widehat{\mathbf{W}}^\top_0 & \widehat{\mathbf{W}}_1^\top & \dots \end{matrix}\right]$.

%\subsubsection{Learning}
%\label{sec:classification_learning}
\begin{flalign}
\textrm{\textbf{Learning:} We optimize $\widehat{\mathbf{W}}$ with:} \hspace{1cm}
\min_{\widehat{\mathbf{W}}} ||\widehat{\mathbf{H}} - \mathbf{Y}||_F^2 
=  \min_{\widehat{\mathbf{W}}} ||\widehat{\mathbf{M}}^{\textrm{(JKN)}}
\widehat{\mathbf{W}} - \mathbf{Y}||_F^2 &&
\label{eq:min_mp}
\end{flalign}
Loss in Equation \ref{eq:min_mp} can perform well on classification tasks, and according to \citet{hui2021squareloss}, as well as the cross entropy loss defined in Equation \ref{eq:xe_mp}. Taking $\nabla_{\widehat{\mathbf{W}}} $ of Eq.~\ref{eq:min_mp} then setting to zero,
\begin{flalign}
\textrm{yields minimizer:} \hspace{1cm}
\widehat{\mathbf{W}}^* \triangleq \argmin_{\widehat{\mathbf{W}}} ||\widehat{\mathbf{M}}^{\textrm{(JKN)}}
\widehat{\mathbf{W}} - \mathbf{Y}||_F^2 = \left(\widehat{\mathbf{M}}^{\textrm{(JKN)}}\right)^\dagger \mathbf{Y}.&&
\end{flalign}
Rank-$k$ SVD can estimate $\left(\widehat{\mathbf{M}}^{\textrm{(JKN)}}\right)^\dagger$ and hence  $\widehat{\mathbf{W}}^*$ as:
\begin{equation}
\widehat{\mathbf{U}}_k, \widehat{\mathbf{S}}_k, \widehat{\mathbf{V}}_k  \leftarrow \textrm{SVD}(\widehat{\mathbf{M}}^{^\textrm{(JKN)}}, k); \hspace{1cm}
\widehat{\mathbf{W}}^*  \leftarrow (\widehat{\mathbf{V}}_k (\widehat{\mathbf{S}}_k^{-1} (\widehat{\mathbf{U}}_k^\top \mathbf{Y}))). \label{eq:w_svd}
\end{equation}
Order of multiplications in Eq.~\ref{eq:w_svd} is for efficiency. Further, in the case when only subset of nodes $\mathcal{V} = \{v\}_v$ have labels, the right-most multiplication of Eq.~\ref{eq:w_svd} could restricted to the labeled nodes. Let $\mathbf{Y}_{\mathcal{V}}$ be a matrix of $|\mathcal{V}|$ rows selected from $\mathbf{Y}$ according to elements $\mathcal{V}$. The right-most multiplication of Eq.~\ref{eq:w_svd} can modified to:
${{\widehat{\mathbf{U}}_{k_\mathcal{V}}}^\top} \mathbf{Y}_\mathcal{V}$.

%In practice, one can get creative in designing $\widehat{\mathbf{M}}^{^\textrm{(JKN)}}$, for instance, extending $\widehat{\mathbf{M}}$  with a column of ones $\mathbf{1}$, to account for offset/bias. For transductive settings, also with powers of transformed adjacency matrix: $\mathbf{I}, g_1(\mathbf{A}), g_2(\mathbf{A})^2, \dots$

\section{Functional Singular Value Decomposition}
\vspace{-0.3cm}
\label{sec:fsvd}
We do not have to explicitly calculate the matrices $\widehat{\mathbf{M}}$. Rather, we only need to implement product functions $f_{\widehat{\mathbf{M}}}(\mathbf{v}) = \langle \widehat{\mathbf{M}}, \mathbf{v} \rangle $ that can multiply $\widehat{\mathbf{M}}$ with arbitrary (appropriately-sized) vector $ \mathbf{v}$. We implement a (TensorFlow) \textbf{functional} version of the randomized  SVD algorithm of \citet{halko2009svd}, that accepts $f_{\widehat{\mathbf{M}}}$ rather than $\widehat{\mathbf{M}}$. We show that it can train our models quickly and with arbitrarily small approximation error (in linear time of graph size, in practice, with less than 10 passes over the data) and can yield l2-regularized solutions for classification (see Appendix). We now need the (straightforward
$f_{\widehat{\mathbf{M}}^{^\textrm{(WYS)}}}$ and $f_{\widehat{\mathbf{M}}^{^\textrm{(JKN)}}}$. We leave the second outside this writing. For the first, the non-edges term, $(1 - \mathbf{A})$, can be re-written by explicit broadcasting as $(\mathbf{1}\mathbf{1}^\top - \mathbf{A})$ giving
\begin{equation}
\label{eq:fmwys}
f_{\widehat{\mathbf{M}}^{^\textrm{(WYS)}}}(\mathbf{v}) = \sum_i      \underbrace{(\mathcal{T})^i \mathbf{v}}_{\mathcal{O}(im)} \mathbf{c}_i  - \lambda \mathbf{1} \underbrace{(\mathbf{1}^\top \mathbf{v})}_{\mathcal{O}(n)} + \lambda \underbrace{\mathbf{A} \mathbf{v}}_{\mathcal{O}(m)}.
\end{equation}
All matrix-vector products can be efficiently computed when $\mathbf{A}$ is sparse.
%For instance, computing $\mathcal{T}^3 \mathbf{v}$ can be calculated right-to-left as $\mathcal{T}(\mathcal{T}(\mathcal{T}( \mathbf{v})))$ using $3m$ floating-point multiplications, consuming no-more than
%$\mathcal{O}(m)$ memory. For space constraints, the other (trivial) implementations for $\widehat{\mathbf{M}}^{^\textrm{(WYS)}}$ and $\widehat{\mathbf{M}}^{^\textrm{(JKN)}}$ are left outside this write up.

\section{Experimental Results (details in Appendix)}
\label{sec:exp}
\vspace{-0.3cm}
\begin{figure}[h]
\includegraphics[width=\linewidth]{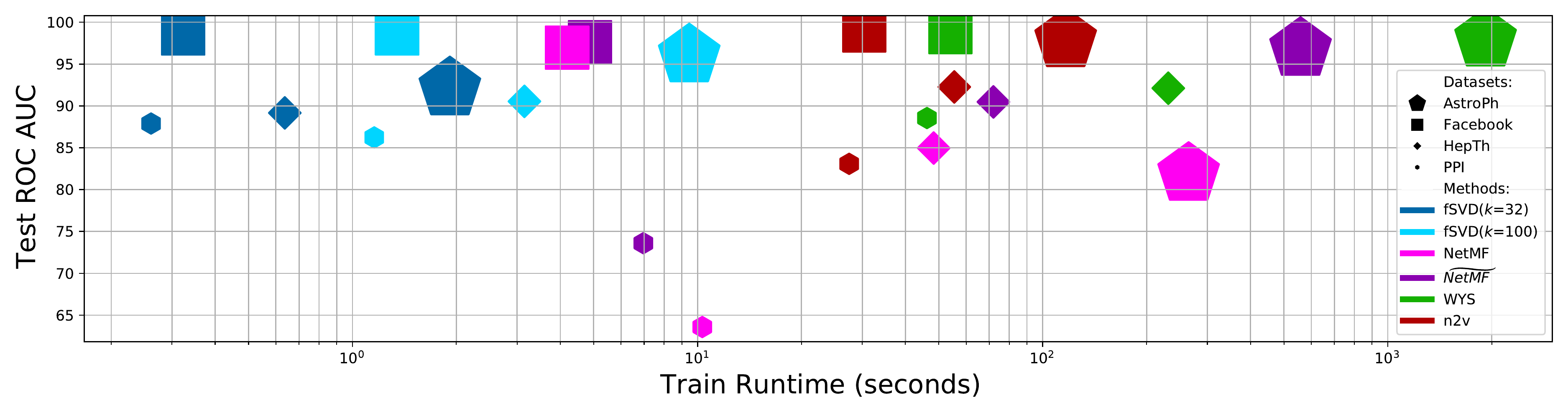}
\vspace{-25pt}
\caption{ROC-AUC versus train time of methods on datasets. Each dataset has a distinct shape, with shape size proportional to graph size.
Each method uses a different color. Our methods are in blue (dark uses SVD rank $k=32$, light uses $k=100$, trading estimation accuracy for train time).
%Focusing on any single shape (e.g., hexagon) comparatively shows performances.
Ideal methods should be placed on top-left corner (i.e., higher test ROC-AUC and faster training).}
\end{figure}
\begin{figure}[h]
\includegraphics[width=0.49\linewidth]{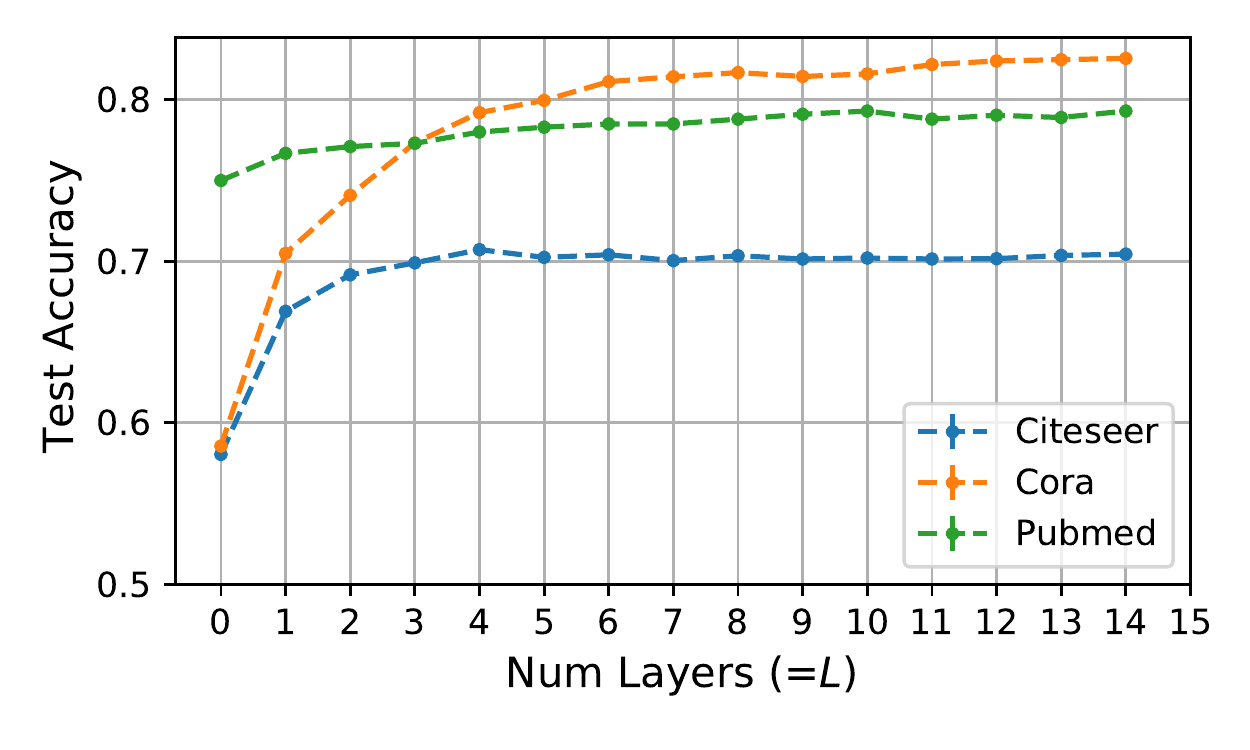}
\includegraphics[width=0.49\linewidth]{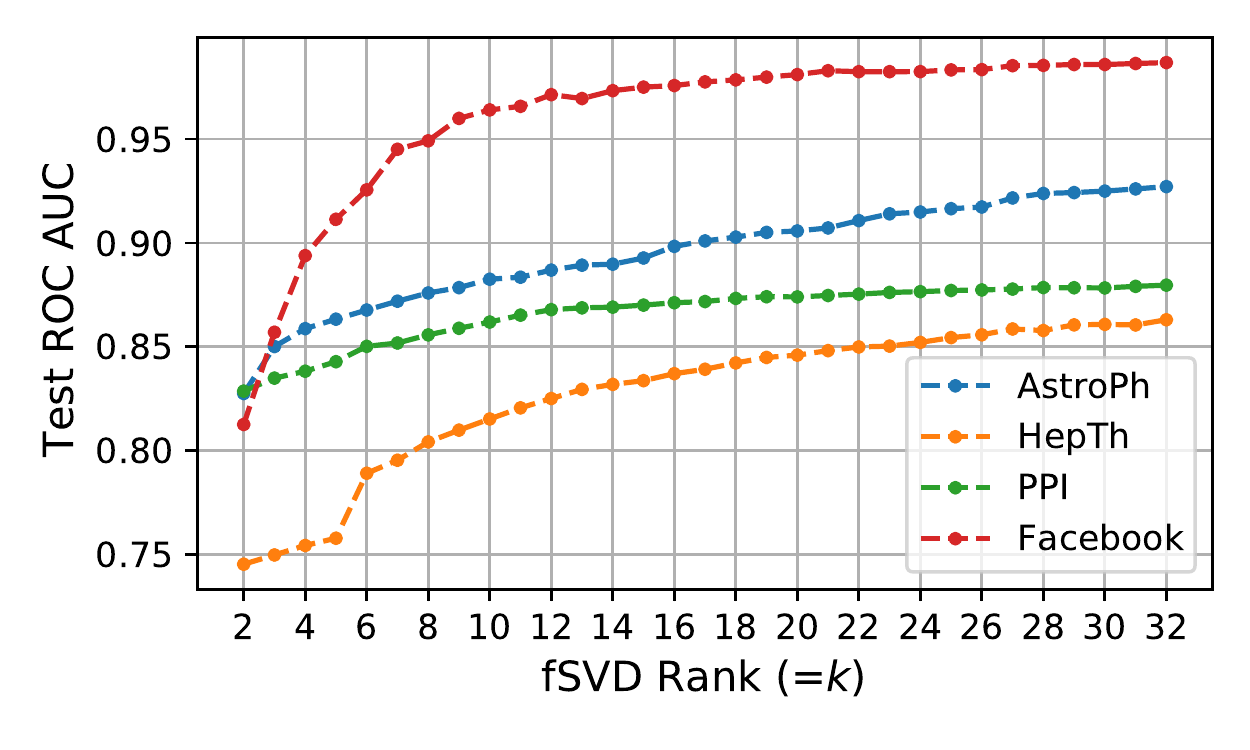}
\vspace{-15pt}
\caption{Sensitivity Analysis. \textbf{Left}: Test Accuracy VS Depth of model defined by $\widehat{\mathbf{M}}^{^\textrm{(JKN)}}$. \textbf{Right}: Test ROC-AUC VS rank of SVD on $\widehat{\mathbf{M}}^{^\textrm{(WYS)}}$.}
\label{fig:sensitivity}
\end{figure}
\begin{table}[h]
\caption{Test Performance. \textbf{Left}: accuracy (training time) for Semi-supervised Node Classification, over citation datasets.
\textbf{Right}:
ROC-AUC for link prediction when embedding with $z$=64=$2k$.}
\label{table:results}
\begin{adjustwidth}{-8pt}{-8pt}
\begin{tabular}{r| l | l| l}
\toprule
%\begin{tabular}{r| cc | cc | cc| cc}
& \multicolumn{1}{c|}{\textbf{Cora}}
& \multicolumn{1}{c|}{\textbf{Citeseer}}
& \multicolumn{1}{c}{\textbf{Pubmed}}
  \\
\hline
Planetoid & 75.7 (13s) & 64.7 (26s) & 77.2 (25s) \\
GCN  &  81.5 (4s)  & 70.3 (7s) & 79.0 (83s) \\
GAT   & 83.2 (1m23s)    & 72.4 (3m27)  & 77.7 (5m33s) \\
MixHop & 81.9 (26s) & 71.4 (31s) & 80.8 (1m16s)  \\
GCNII & 85.5 (2m29s)  & 73.4 (2m55s)  & 80.3 (1m42s) \\
\hline
\hline
$f_{\widehat{\mathbf{M}}^\textrm{(JKN)}}$ & 82.4 (0.28s)  & 72.2 (0.13s)  & 79.7 (0.14s) \\
% \hline
\bottomrule
\end{tabular}{}
%\caption{}
%\label{table:wys}
\hspace{5pt}
\setlength\tabcolsep{2pt} % default value: 6pt
\begin{tabular}{r| c | c | c | c}
\toprule
%\begin{tabular}{r| cc | cc | cc| cc}
& \multicolumn{1}{c|}{\textbf{FB}}
& \multicolumn{1}{c|}{\textbf{AstroPh}}
& \multicolumn{1}{c|}{\textbf{HepTh}}
& \multicolumn{1}{c}{\textbf{PPI}}
  \\
\hline
WYS & 99.4  & 97.9  & 93.6  & 89.8  \\
n2v & 99.0  & 97.8  & 92.3  & 83.1  \\
NetMF & 97.6  & 96.8  & 90.5  & 73.6  \\
$\widetilde{\textrm{NetMF}}$ & 97.0  & 81.9  & 85.0  & 63.6  \\
\hline
\hline
$f_{\widehat{\mathbf{M}}^\textrm{(WYS)}}$ & 98.7  & 92.1 & 89.2  & 87.9  \\
($k$=100)  & 98.7  & 96.0 & 90.5 & 86.2 \\
% \hline
\bottomrule
\end{tabular}{}
\end{adjustwidth}
\end{table}

\newpage

\subsubsection*{Acknowledgments}
This material is based upon work supported by the Defense Advanced
Research Projects Agency (DARPA) and the Army Contracting Command-Aberdeen Proving Grounds (ACC-APG) under Contract Number
W911NF-18-C-0020.

%\subsubsection*{Author Contributions}
%If you'd like to, you may include  a section for author contributions as is done
%in many journals. This is optional and at the discretion of the authors.

%\subsubsection*{Acknowledgments}
%Use unnumbered third level headings for the acknowledgments. All
%acknowledgments, including those to funding agencies, go at the end of the paper.

\bibliography{iclr2021_workshop}
\bibliographystyle{iclr2021_workshop}

\appendix
\section{Appendix: Implementation of Functional SVD}

\subsection{Calculating the SVD}
SVD of  $\mathbf{M}$ yields its \textbf{left} and \textbf{right} singular orthonormal (basis) vectors, respectively, in columns of $\mathbf{U}$ and $\mathbf{V}$. Since $\mathbf{U}$ and $\mathbf{V}$, respectively, are the eigenvectors of $\mathbf{M} \mathbf{M}^\top$ and $\mathbf{M}^\top \mathbf{M}$,
then perhaps the most intuitive algorithms for SVD are variants of the \textit{power iteration}, including Arnoldi iteration \citep{arnoldi} and Lanczos algorithm \citep{lanczos}.
In practice, randomized algorithms for estimating SVD run faster than these variants, including the algorithm of \citet{halko2009svd} which is implemented in scikit-learn. \underline{None} of these methods require individual access to $\mathbf{M}$'s entries, but rather, require two operations: ability to multiply any vector with $\mathbf{M}$ and with $\mathbf{M}^\top$. Therefore, it is only a practical gap that we fill in this section: we open-source a TensorFlow implementation that accept product and transpose operators.

\subsection{TensorFlow Implementation} 
Since we do not explicitly calculate the $\widehat{\mathbf{M}}$ matrices displayed in Equations \ref{eq:m_wys} and \ref{eq:h_jkn}, as doing so consumes quadratic memory $\mathcal{O}(n^2)$, we implement  a functional form SVD of the celebrated randomized SVD algorithm of \citet{halko2009svd}.
To run our Algorithm \ref{alg:fsvd}, one must specify $k \in \mathbb{N}_+$ (rank of decomposition), as well as functions $f, l, s$ that the program provider promises they operate as:
\begin{enumerate}[itemsep=0pt, topsep=0pt, leftmargin=12pt]
\item Product function $f$ that exactly computes $f(\mathbf{v}) = \langle \mathbf{M} , \mathbf{v} \rangle$ for any $\mathbf{v} \in \mathbb{R}^c $  (recall: $\mathbf{M} \in \mathbb{R}^{r \times c}$)
\item Transpose\footnote{An alternative to $t$ can be a left-multiply function $l(\mathbf{u}) = \langle \mathbf{u}, \mathbf{M}  \rangle$, however, in practice, TensorFlow is optimized for CSR matrices and computationally favors sparse-times-dense} function $t$. $\forall \mathbf{v} \in \mathbb{R}^r$,  $(t \circ f)(\mathbf{v}) = \langle \mathbf{M}^\top, \mathbf{v} \rangle $% for $$
%Structured transpose is necessary, as
%we do not explicitly calculate $\mathbf{M}$.
\item Shape (constant) function $s$ that knows and returns $(r, c)$. Once transposes, should return $(c, r)$.
\end{enumerate}

\definecolor{CommentColor}{HTML}{006619}
\algrenewcommand\algorithmicindent{1.0em}%
\begin{algorithm}[h]
   \caption{Functional Randomized SVD, following prototype of \citet{halko2009svd}}
   \label{alg:fsvd}
\begin{algorithmic}[1]
   \State {\bfseries input:} rank $k \in \mathbb{N}_+$, product fn $f : \mathbb{R}^c \rightarrow \mathbb{R}^r$, shape fn $s$,  transpose fn $t : (\mathbb{R}^c \rightarrow \mathbb{R}^r) \rightarrow (\mathbb{R}^c \rightarrow \mathbb{R}^r)$
   \Procedure{\textnormal fSVD}{$f, t, s, k$}
   \State $(r, c) \leftarrow s()$
   \State $Q \sim \mathcal{N}(0, 1)^{c \times 2k}$
   \Comment{\textcolor{CommentColor}{IID Gaussian. Shape: $(c \times 2k)$}}
   \For{$i \leftarrow 1$ {\bfseries to} \texttt{iterations}}
   \State $Q, \_ \leftarrow \texttt{tf.linalg.qr}(f(Q))$
   \Comment{\textcolor{CommentColor}{$(r \times 2k)$}}
   \State $Q, \_ \leftarrow \texttt{tf.linalg.qr}((t \circ f)(Q))$
   \Comment{\textcolor{CommentColor}{$(c \times 2k)$}}
   \EndFor
   \State $Q, \_ \leftarrow \texttt{tf.linalg.qr}(f(Q))$
   \Comment{\textcolor{CommentColor}{$(r \times 2k)$}}
   \State $B \leftarrow ((t \circ f)(Q))^\top$
   \Comment{\textcolor{CommentColor}{$(2k \times c)$}}
   \State $U, s, V^\top \leftarrow \texttt{tf.linalg.svd}(B)$
   \State $U \leftarrow Q \times U$
   \Comment{\textcolor{CommentColor}{$(r \times 2k)$}}
   \State \textbf{return } $U[:, :k], s[:k], V[:, :k]^\top$
   \EndProcedure
\end{algorithmic}
\end{algorithm}

\subsection{Analysis}
\subsubsection{Norm Regularization of Wide Models}
\label{sec:overfitting}
%In a sense, SVDs inherit a benefit.
If $\widehat{\mathbf{M}}$ is too wide,
%(e.g., contains transformed adjacencies),
%Then here are 
%If $k$ was set equal
then we need \textbf{not} to worry much about \textit{overfitting},
%, e.g. if, so long that $k$ is small-enough to not memorize (perfectly reconstruct) the input data,
due to the following Theorem.
%\citep{eckart1936lowrank}
%, which states that: \newline
\begin{theorem}
\label{theorem:min_norm}
{\normalfont (Min. Norm)}
If system $\widehat{\mathbf{M}} \widehat{\mathbf{W}} = \mathbf{Y}$ is underdetermined\footnote{E.g., if the number of labeled examples i.e. height of $\mathbf{M}$ and $\mathbf{Y}$ is smaller than the width of $\mathbf{M}$.} with rows of $\widehat{\mathbf{M}}$ being linearly independent, then there are infinitely many solutions. Denote solution space $\widehat{\mathcal{W}}^* = \Big\{\widehat{\mathbf{W}} \ \Big| \  \widehat{\mathbf{M}} \widehat{\mathbf{W}} = \mathbf{Y} \Big\}$.
%\citep{eckart1936lowrank}: \textit{
%
Then, for $k\geq \textrm{rank}(\widehat{\mathbf{M}})$, matrix $\widehat{\mathbf{W}}^*$, defined in Eq.\ref{eq:w_svd} satisfies:
$
\widehat{\mathbf{W}}^* = \mathop{\mathrm{argmin}}_{\widehat{\mathbf{W}}\in \widehat{\mathcal{W}}^* } ||\widehat{\mathbf{W}}||_F^2 
$
\end{theorem}
%%%%%%%%%%%%%%% PROOF OF \ref{theorem:min_norm}
\textbf{Proof} Assume $\mathbf{Y} =\mathbf{y}$ is a column vector (the proof can be generalized to matrix $\mathbf{Y}$ by repeated column-wise application\footnote{The minimizer for the Frobenius norm is composed, column-wise, of the minimizers $\mathop{\mathrm{argmin}}_{\widehat{\mathbf{M}} \mathbf{W}_{:, j} = \mathbf{Y}_{:, j}} ||\mathbf{W}_{:, j}||_2^2$ for all $j$.}). SVD$(\widehat{\mathbf{M}}, k)$, $k\geq \textrm{rank}(\widehat{\mathbf{M}})$, recovers the solution:
\begin{equation}
\widehat{\mathbf{W}}^* = \left(\widehat{\mathbf{M}}\right)^\dagger \mathbf{y} = \widehat{\mathbf{M}}^\top \left(\widehat{\mathbf{M}} \widehat{\mathbf{M}}^\top \right)^{-1} \mathbf{y}.
\end{equation}
The \textit{Gram matrix} $\widehat{\mathbf{M}} \widehat{\mathbf{M}}^\top$ is nonsingular as the rows of $\widehat{\mathbf{M}}$ are linearly independent.
To prove the claim let us first verify that $\widehat{\mathbf{W}}^* \in \widehat{\mathcal{W}}^*$:
\begin{equation*}
\widehat{\mathbf{M}} \widehat{\mathbf{W}}^*
= \widehat{\mathbf{M}} \widehat{\mathbf{M}}^\top \left(\widehat{\mathbf{M}} \widehat{\mathbf{M}}^\top \right)^{-1} \mathbf{y} = \mathbf{y}.
\end{equation*}
Let $\widehat{\mathbf{W}}_p  \in \widehat{\mathcal{W}}^*$.
We must show that $||\widehat{\mathbf{W}}^*||_2 \le ||\widehat{\mathbf{W}}_p||_2$.
Since  $\widehat{\mathbf{M}} \widehat{\mathbf{W}}_p = \mathbf{y}$ and $\widehat{\mathbf{M}} \widehat{\mathbf{W}}^* = \mathbf{y}$, their subtraction gives:
\begin{equation}
\label{eq:nullM}
\widehat{\mathbf{M}} ( \widehat{\mathbf{W}}_p - \widehat{\mathbf{W}}^* ) = 0.
\end{equation}
It follows that $( \widehat{\mathbf{W}}_p - \widehat{\mathbf{W}}^* )  \perp \widehat{\mathbf{W}}^*$:
\begin{align*}
    ( \widehat{\mathbf{W}}_p - \widehat{\mathbf{W}}^* )^\top  \widehat{\mathbf{W}}^* &= ( \widehat{\mathbf{W}}_p - \widehat{\mathbf{W}}^* )^\top \widehat{\mathbf{M}}^\top \left(\widehat{\mathbf{M}} \widehat{\mathbf{M}}^\top \right)^{-1} \mathbf{y} \\
    &= \underbrace{(\widehat{\mathbf{M}} (\widehat{\mathbf{W}}_p - \widehat{\mathbf{W}}^* ))^\top}_{=0 \textrm{ due to Eq.~\ref{eq:nullM}}} \left(\widehat{\mathbf{M}} \widehat{\mathbf{M}}^\top \right)^{-1} \mathbf{y} = 0
\end{align*}
Finally, using Pythagoras Theorem (due to $\perp$):
% ||\widehat{\mathbf{W}}^*||_2 \le ||\widehat{\mathbf{W}}_p||_2
\begin{align*}
||\widehat{\mathbf{W}}_p||_2^2 &= ||\widehat{\mathbf{W}}^* + \widehat{\mathbf{W}}_p - \widehat{\mathbf{W}}^*||_2^2 \\
&= ||\widehat{\mathbf{W}}^*||_2^2 + ||\widehat{\mathbf{W}}_p - \widehat{\mathbf{W}}^*||_2^2 \ge ||\widehat{\mathbf{W}}^*||_2^2
\ \ \ \ \ \ \ \null\hfill \blacksquare
\end{align*}    
%%%%%%%%%%%%%%% end of PROOF OF \ref{theorem:min_norm}

As a consequence, solution for classification models recovered by SVD follow a strong standard Gaussian prior, which may be regarded as a form of regularization.
%on the solution $\widehat{\mathbf{W}}$ calculated in Equation \ref{eq:w_svd}. 

%\section{Application on Graphs}
%WRITE ME 

\subsubsection{Computational Complexity and Approximation Error}
\label{sec:analysis}
\begin{theorem}
\label{theorem:linear_time}
{\normalfont (Linear Time)}
%(Linear Time)
Functional SVD (Alg.~\ref{alg:fsvd}) trains our convexified GRL models in time linear in the graph size.
\end{theorem}
%\newline
\textbf{Proof} of Theorem \ref{theorem:linear_time} for our two model families:
\begin{enumerate}[itemsep=0pt,topsep=0pt,leftmargin=12pt]
\item For rank-$k$ SVD over $f_{\widehat{\mathbf{M}}^\textrm{(WYS)}}$:
Let cost of running $f_{\widehat{\mathbf{M}}^\textrm{(WYS)}} = T_\textrm{mult}$. The run-time to compute SVD, as derived in Section 1.4.2 of \citep{halko2009svd}, is:
\begin{equation}
\mathcal{O}(k T_\textrm{mult} + (r + c) k^2).   
\end{equation}
Since $f_{\widehat{\mathbf{M}}^\textrm{(WYS)}}$ can be defined as $C$ (context window size) multiplications with sparse $n \times n$ matrix $\mathcal{T}$ with $m$ non-zero entries, then %$T_\textrm{mult}=mC$ and
running fSVD($f_{\widehat{\mathbf{M}}^\textrm{(WYS)}}, k$) costs:
\begin{equation}
\mathcal{O}(k m C + n  k^2)
\end{equation}
\item For rank-$k$ SVD over $f_{\widehat{\mathbf{M}}^\textrm{(JKN)}}$: Suppose feature matrix contains $d$-dimensional rows. One can calculate $\widehat{\mathbf{M}}^\textrm{(JKN)} \in \mathbb{R}^{n \times L d}$ with $L$ sparse multiplies in $\mathcal{O}(L m d)$. Calculating and running SVD \citep[see Section 1.4.1 of][]{halko2009svd} on $\widehat{\mathbf{M}}^\textrm{(JKN)}$ costs total of:
\begin{equation}
    \mathcal{O}(n d L \log(k) + (n + d L ) k ^ 2 + L m d).
\end{equation}
\end{enumerate}
Therefore, training time is linear in $n$ and  $m$. $\ \  \blacksquare$

Contrast with methods of WYS \citep{wys} and NetMF \citep{qiu2018network}, which require assembling a dense $n \times n$ matrix  requiring $\mathcal{O}(n^2)$ time to decompose. One wonders: how far are we from the optimal SVD with a linear-time algorithm? The following bounds the error.

\begin{theorem}
\label{theorem:error}
{\normalfont (Exponentially-decaying Approx. Error)}
Rank-$k$  randomized SVD algorithm of \citet{halko2009svd} gives an approximation error that can be brought down, exponentially-fast, to no more than twice of the approximation error of the optimal (true) SVD.
\end{theorem}
\textbf{Proof} is in Theorem 1.2 of \citet{halko2009svd}. $\ \  \blacksquare$
\newline
Consequently,
compared to $\widetilde{\textrm{NetMF}}$ of \citep{qiu2018network}, which incurs unnecessary estimation error,
our estimation error can be brought-down exponentially by increasing the \texttt{iters} parameter of Alg.~\ref{alg:fsvd}  %the  negligable error bounds 
%\subsection{Error Bounds}
%WRITE ME 

\section{Appendix: Experimental Details}

\subsection{Datasets}
\label{sec:datasets}
We apply our functional SVD  on popular datasets that can be trained using our simplified (i.e., convexified) models. Specifically either (1) semi-supervised node-classification datasets, where features are present, or (2) link-prediction datasets where features are absent. It is possible to convexify other setups, e.g., link prediction when node features are present, but we leave this as future work. We run experiments on
seven graph datasets:
\begin{itemize}[itemsep=0pt, topsep=0pt]
\item Protein-Protein Interactions (PPI): a large graph where every node is a protein and an edge between two nodes indicate that the two proteins interact. Processed version of PPI was downloaded from \citep{grover2016node2vec}.
\item Three citation networks that are extremely popular: Cora, Citeseer, Pubmed. Each node is an article and each (directed) edge implies that an article cites another. Additionally, each article is accompanied with a feature vector (containing NLP-extracted features of the article's abstract), as well as a label (article type). %The names of these datasets indicate the source of these articles 
\item Two collaboration datasets: ca-AstroPh and ca-HepTh, where nodes are researchers and an edge between two nodes indicate that the researchers co-published together at least one article, in the areas Astro-Physics and High Energy Physics.
\item ego-Facebook: an ego-centered social network.
\end{itemize}

For citation networks, we processed node features and labels. For all other datasets, we did not process features during training nor inference.
For train/validation/test partitions: we used the splits of \citet{planetoid} for Citeseer, Cora, Pubmed; we used the splits of \citet{wys} for PPI, Facebook, ca-AstroPh and ca-HepTh; we used the splits of OGB \citep{ogb} for ogbl-ddi.
All datasets and statistics are summarized in Table \ref{table:datasets}.
In §\ref{sec:exp_classification} and §\ref{sec:exp_lp}, unless otherwise noted, we download authors' source code from github, modify\footnote{Modified files are in our code repo} it to record wall-clock run-time, and run on GPU NVidia Tesla k80. Thankfully, downloaded code has one script to run each dataset, or hyperparameters are clearly stated in the source paper.

\setlength\tabcolsep{4pt} % default value: 6pt
\begin{table*}
\caption{Dataset Statistics}
\label{table:datasets}
\begin{adjustwidth}{-5pt}{-5pt}
\begin{tabular}{r| r  l | r l | l}
\toprule
\textbf{Dataset} & \multicolumn{2}{c|}{\textbf{Nodes}} & \multicolumn{2}{c|}{\textbf{Edges}} & \multicolumn{1}{c}{\textbf{Source}} \\
\midrule
PPI & 3,852 & proteins & 20,881 & chemical interactions & {\small \href{http://snap.stanford.edu/node2vec}{http://snap.stanford.edu/node2vec}} \\
ego-Facebook & 4,039  & users & 88,234 & friendships & {\small
\href{http://snap.stanford.edu/data}{http://snap.stanford.edu/data}} \\
ca-AstroPh & 17,903 & researchers & 197,031 & co-authorships & {\small
\href{http://snap.stanford.edu/data}{http://snap.stanford.edu/data}} \\
ca-HepTh & 8,638 & researchers & 24,827 & co-authorships & {\small
\href{http://snap.stanford.edu/data}{http://snap.stanford.edu/data}} \\
Cora & 2,708 & articles & 5,429 & citations & Planetoid \citep{planetoid} \\
Citeseer &  3,327 &articles &  4,732 & citations & Planetoid \citep{planetoid}\\
Pubmed & 19,717 & articles & 44,338 & citations & Planetoid \citep{planetoid} \\
\bottomrule
\end{tabular}
\end{adjustwidth}
\end{table*}

\subsection{Semi-supervised Node Classification}
\label{sec:exp_classification}
We consider a \textit{transductive} setting where a graph is entirely visible (all nodes and edges). Additionally, some of the nodes are labeled. The goal is to recover the labels of unlabeled nodes. All nodes have feature vectors. 

\textbf{Baselines}: We download code of GAT \citep{gat}, MixHop \citep{mixhop}, GCNII \citep{GCNII} and re-ran them, with slight modifications to record training time. However, for baselines Planetoid \citep{planetoid} and GCN \citep{kipf}, we copied them from the GCN paper \citep{kipf}.

In these experiments, to train our method, we run our functional SVD twice per graph.
We take the feature matrix $\mathbf{X}$ bundled with the datasets, and concatenate to it two matrices, $\widehat{\mathbf{L}}$ and $\widehat{\mathbf{R}}$, calculated per Equation \ref{eq:compute_lr}: the calculation itself invokes our functional SVD (the first time) on $f_{\widehat{\mathbf{M}}^\textrm{(WYS)}}$ with rank $=32$.
Hyperparameters of $f_{\widehat{\mathbf{M}}^\textrm{(WYS)}}$ are $\lambda$ (negative coefficient) and $C$ (context window-size). After concatenating $\widehat{\mathbf{L}}$ and $\widehat{\mathbf{R}}$ into $\mathbf{X}$, we PCA the resulting matrix to 1000 dimensions, which forms our new $\mathbf{X}$. Finally, we express our model as the linear $L$-layer messaging passing network (Eq.~\ref{eq:h_jkn}) and learn its parameters via rank $k$ SVD on $f_{\widehat{\mathbf{M}}^{\textrm{(JKN)}}}$ (the second time), as explained in §\ref{sec:classification_learning}.
We use the validation partition to tune 
$L$, $k$, $\lambda$, and $C$.

Table \ref{table:results} (left) summarizes the performance of our approach ($f_{\widehat{\mathbf{M}}^\textrm{(JKN)}}$) against aforementioned baselines, showing both test accuracy and training time. While our method is competitive with state-of-the-art, it trains much faster.

\subsection{ROC-AUC Link Prediction}
\label{sec:exp_lp}
Given a partial graph: only of a subset of edges are observed. The goal is to recover unobserved edges.
This has applications in recommender systems: when a user expresses interest in products, the system wants to predict other products the user is interested in. The task is usually setup by partitioning the edges of the input graph into train and test edges. Further, it is common to sample \textit{negative test edges} e.g. uniformly from the graph compliment. Lastly, a GRL method for link prediction can be trained on the train edges partition, then can be asked to score the test partition edges versus the \textit{negative test edges}. The quality of the scoring can be quantified by a ranking metric, e.g., ROC-AUC.

\textbf{Baselines}: We download code of WYS \citep{wys}
and update it to for TensorFlow-2.0.
We download code of \citet{qiu2018network} and denote their
methods as NetMF and $\widetilde{\textrm{NetMF}}$, where the first computes complete matrix $\mathbf{M}$ before SVD decomposition and the second sample $\mathbf{M}$ entry-wise -- the second is faster for larger graphs but sacrifices on estimation error and performance.
For node2vec (n2v), we use its PyG implementation \citep{pyg}.

Table \ref{table:results} (right) summarizes results test ROC-AUC.
For our method (denoted $f_{\widehat{\mathbf{M}}^\textrm{(WYS)}}$), we call our functional SVD (Alg.~\ref{alg:fsvd}) and pass it $f_{\widehat{\mathbf{M}}^\textrm{(WYS)}}$ as defined in Eq.~\ref{eq:fmwys}.
%, setting $\lambda=3$.
Embeddings are set to the SVD basis (as in, §\ref{sec:embeddingmodel}) and edge score of nodes $(u, v)$ is $\propto$ dot-product of embeddings. The last row of the table shows results when svd rank $=$100.
Lastly, 
we set the context window hyperparameter (a.k.a, length of walk) as follows. For WYS, we trained with their default context (as WYS learns the context), but for all others (NetMF, n2v, ours) we used context window of length $C$=5 for datasets Facebook and PPI (for us, this sets $\mathbf{c} = [5, 4, 3, 2, 1]$) and $C$=20 for AstroPh and HepTh.

\subsection{Sensitivity Analysis}
While in §\ref{sec:exp_classification} we tune the number of layers ($L$) using the performance on the  validation partition, in this section, we show impact of varying $L$ on test accuracy.
According to the summary in Figure~\ref{fig:sensitivity} (left), accuracy of classifying a node improves when incorporating information from further nodes. We see little gains beyond $L>6$. Note that $L=0$ corresponds to ignoring the adjacency matrix altogether when running $f_{\widehat{\mathbf{M}}^\textrm{(JKN)}}$. Here, we fixed $\lambda = C = 1$ and averaged 5 runs. The (tiny) error bars show the standard deviation.

Further, while in §\ref{sec:exp_lp} we do SVD on $f_{\widehat{\mathbf{M}}^\textrm{(WYS)}}$ with rank $k=32$ or $k=100$, Figure~\ref{fig:sensitivity} (right) shows test accuracy while sweeping $k \le 32$. In general, increasing the rank improves estimation accuracy and test performance. However, if $k$ is larger than the inherit dimensionality of the data,
%(can be $<n$, due to the \textit{Manifold Hypothesis}),
then this could cause overfitting (though perfect memorization of training edges).
The \textit{Norm Regularization} note (§\ref{sec:overfitting}) applies only to pseudoinversion i.e.  our classification models.

%\section{Experiments}
%\label{sec:exp}

\end{document}

%% file: math_commands.tex
\usepackage{amsmath,amsfonts,bm}

\def\eqref#1{equation~\ref{#1}}
\def\1{\bm{1}}

\DeclareMathAlphabet{\mathsfit}{\encodingdefault}{\sfdefault}{m}{sl}
\SetMathAlphabet{\mathsfit}{bold}{\encodingdefault}{\sfdefault}{bx}{n}

\DeclareMathOperator*{\argmin}{arg\,min}